# Content-Based Bird Retrieval using Shape context, Color moments and Bag of Features

Bahri abdelkhalak [1] and hamid zouaki [2]

[1]Faculty of Sciences, University Chouaib Doukkali,
Equipe: Modélisation mathématique et informatique décisionnelle
El Jadida, Morocco

[2]Faculty of Sciences, University Chouaib Doukkali,
Equipe: Modélisation mathématique et informatique décisionnelle
El Jadida, Morocco

**Abstract**
In this paper we propose a new descriptor for birds search. First, our work was carried on the choice of a descriptor. This choice is usually driven by the application requirements such as robustness to noise, stability with respect to bias, the invariance to geometrical transformations or tolerance to occlusions. In this context, we introduce a descriptor which combines the shape and color descriptors to have an effectiveness description of birds. The proposed descriptor is an adaptation of a descriptor based on the contours defined in article Belongie et al. [5] combined with color moments [19]. Specifically, points of interest are extracted from each image and information's in the region in the vicinity of these points are represented by descriptors of shape context concatenated with color moments. Thus, the approach bag of visual words is applied to the latter.
The experimental results show the effectiveness of our descriptor for the bird search by content.

## 1. Introduction

Image search in general and CBIR in particular are the search fields in of information management which a large number of methods have been proposed and studied but generally satisfactory solutions do not exist. The need for adequate solutions is growing due to the increasing amount of images produced digitally in various fields, which requires new ways to access the images. In this paper we discuss a special case of image search by content such as content-based retrieval of birds. The choice of this type of images comes to diversity and the large number of birds exists in the world. For effective research, an important criterion comes into play based on the quality of description. In this paper we combined the shape and the color descriptors. The choice of these descriptors is motivated by the nature of the type of images search. Indeed, we find two birds have the same shape, but they are completely are different in color. For this reason we proposed to combine both color and shape to have an effective description. For the shape descriptor we choose to use the shape context [5]. The choice of this descriptor is valorized by its effectiveness in several applications [16, 17].

For the color descriptor, we use the color moments [19], this descriptor have been combined with different descriptor to produce efficient description [7, 18,20].

To search similar birds, a similarity measure between descriptors must be defined. However, the comparison between two birds is not always easy and sometimes it requires some invariance transformations. In addition, the size of the descriptor is often high, which increases the complexity during the search similar objects in a large collection. To do this, the descriptors are grouped into classes to construct a visual vocabulary [4], which results in a process of abstraction. Each class is considered a visual word, and the shape contexts [5] combined with the color moments belonging to the same class share similar informations regardless of the image where the point of interest is extracted. Finally, each bird is described by visual words and paired with the bird query.

Several methods have been proposed to extract feature characteristics of birds, for example Lin et al. [1] proposes a shape ontology framework which integrates visual and domain information, applied to bird classification based on both domain and visual knowledge. In [2,3], the authors use of automated methods for bird species identification.

This paper proposes a descriptors which integrates shape context [5] and color moments applied to bird search based on the BOF[4] approaches. This paper is organized as follows. In Section 2, we present the shape context descriptor to describe a bird. Our contribution to indexing birds is discussed in Section 3. Then, we present an evaluation of our approach in section 4. Finally, in Section 5, we conclude and give perspectives to our work.

## 2. Background of shape context

The shape context to a contour point pi of a shape is determined by the distribution of contour points in the region in the vicinity of $p_i$ [5]. Is a histogram of the relative coordinates of contour points with respect to $p_i$ points that are the reference points.

One shape is represented by the set of sampled points on external and internal boundaries, $C = \{p_1, p_2, .. p_n\}$, $p_i \in R^2$ where n is the number of contour points. For a point pi, the relative coordinates of the n-1 other points are determined. The coordinates are the coordinates of the point in a log-polar coordinate systems using $p_i$ as the origin:

$$q = (\log(r_q), \theta_q), \forall q \neq p_i \wedge q \in C \quad (1)$$

where $r_q$ is the distance between $p_i$ and $q$, $\theta_q$ is the angle between the vector $\overrightarrow{p_i q}$ and the horizontal axis. The shape context $h_i$ of point $p_i$ is defined as:

$$h_i(l) = nomber\{q \neq p_i : (q - p_i) \in bin(l)\}, l = \overline{1, L} \quad (2)$$

Where $h_i(l)$ is the number of contour points belonging to the $l^{th}$ class of the histogram and $bin(l) = \{(r_q, \theta_q): r_q \in [r_l, r_l + \Delta_{rl}] \wedge \theta_q \in [\theta_l, \theta_l + \Delta_{\theta l}]\}$. Shape of the object O is described as the set of shape contexts of contour points:

$$O = \{h_i \mid p_i \in C\} \quad (3)$$

However, the shape context described above is not invariant to rotation and scaling. For invariance to scaling, the radial distances are normalized by the average distance α of $n^2$ pairs of points of the shape [5]. The authors also suggested using the tangent vector associated with each point rather than the absolute horizontal axis for that the shape context is invariant to rotation.

In the literature a description method (CFPI[6]) is proposed. This method is an adaptation of the shape context descriptor [5] to graphic symbols. Specifically, points of interest are extracted from each symbol and information in the region in the vicinity of these points is represented by shape context descriptors.

In our case, we combine the method proposed in [6] with we adding the color information's of each region, then we apply the BOF[4] to them. The details of our approach are presented in the next section.

## 3. Proposed descriptor

According to [5], the distance between two shapes is measured as the sum of the costs of symmetrical best matches shape contexts. This causes a problem of complexity when trying shapes (objects) among many similar candidates. To reduce the complexity of matching calculation between objects, the approach bag of visual words [4] will be operated. The details of our approach will be described as follow:

In the first step we detect the interest points for the bird. The researches described in [8, 9, 10, 11] showed that an object can be efficiently located from its points interest. There are many methods proposed to detect interest points in an image [12, 13, 14]. We chose the DoG (Difference-of-Gaussian) detector presented in [8] for our experiments.

In the second step, we calculate for each interest point it shape context. Each point of them is considered as reference point to calculate it corresponding shape context. So that the object is well represented regardless of its orientation and size, the descriptor should be invariant to rotation and scale, the relative coordinates of the contour points should be normalized.

For an effective description of each interest region, the Color Moments are used to extract the color features from the region of 5x5 pixels around the interest point. Since most of the Color information's are concentrated on the low order moments, only the first moment (mean) and the second moments (variance) will be used as the color features.

In the final step, the Algorithm to build visual words [4] is applied on the set of shape contexts concatenate with first and second color moments to build a bag of visual word that are include both the shape and color information's.

## 4. Experiment results

Let us first introduce the bird dataset we used in our experiments and the performance evaluation measures. We will then present the experimental results.

4.1 The Birds Dataset and Performance Evaluation

To conduct the experimental results, we will focus on the CUB-200[15] dataset. This dataset contains 11788 images of 200 bird species. In our experiments we choose 30 images for each the 200 bird species.

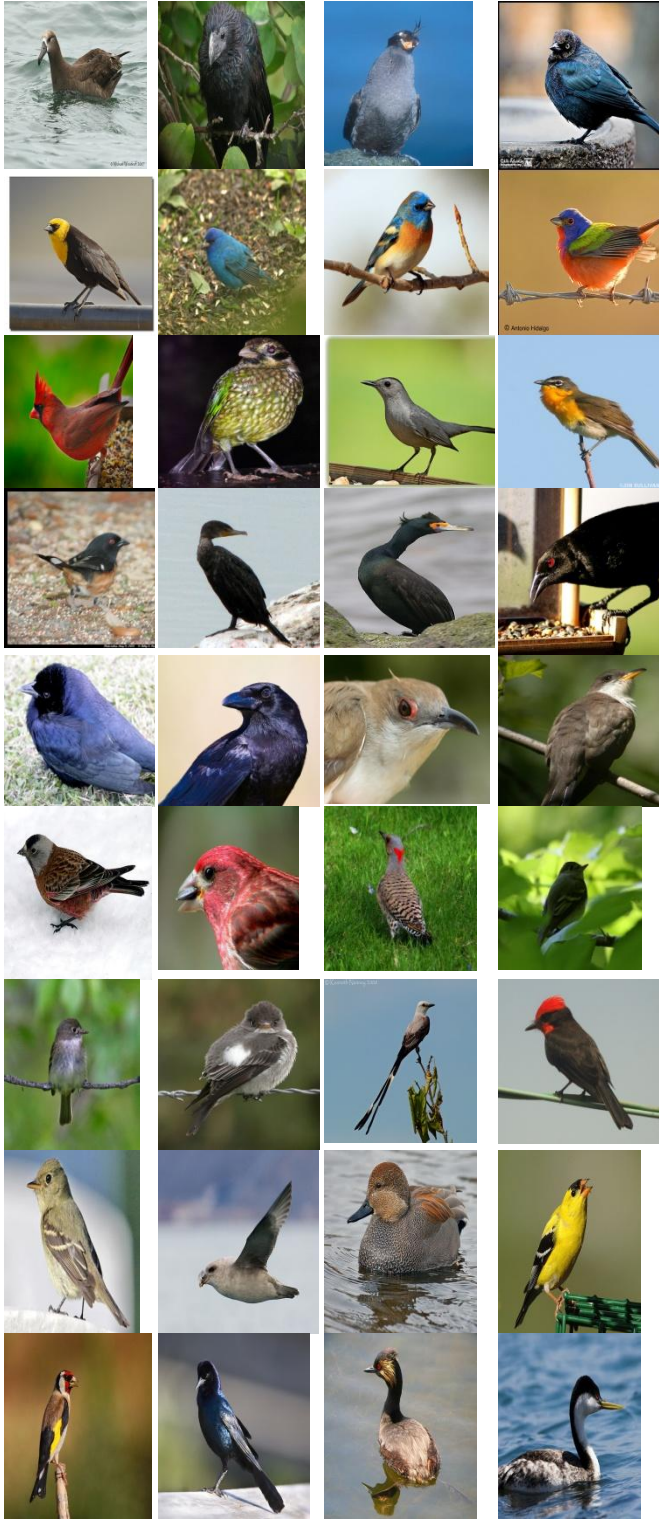

Fig. 1 Sample of WANG Image Database

### 4.2 Results

Before indexing the birds by our method, we computed the set of shape contexts on a database of birds. We use the k-means algorithm to define the visual vocabulary, and then we indexed each bird according to this vocabulary.

In the second training process, Shape context, the first, and the second color moments of each interest points are extracted from each bird in the database, and then are concatenated. After the K-means clustering is used to build another visual vocabulary. In the end, each bird of the database is index according the new vocabulary.

To check the performance of our proposed approach the precision and recall is used. The standard definitions of these two measures are given by following equations.

$$\text{Precision} = \frac{Number\ of\ relevant\ images\ retrieved}{Number\ of\ images\ retrieved} \quad (4)$$

$$\text{Recall} = \frac{Number\ of\ relevant\ images\ retrieved}{Total\ number\ of\ relevant\ images\ in\ the\ database} \quad (5)$$

Let us first take a look at the qualitative results. We can see in figure 2 some precision/recall curve results.

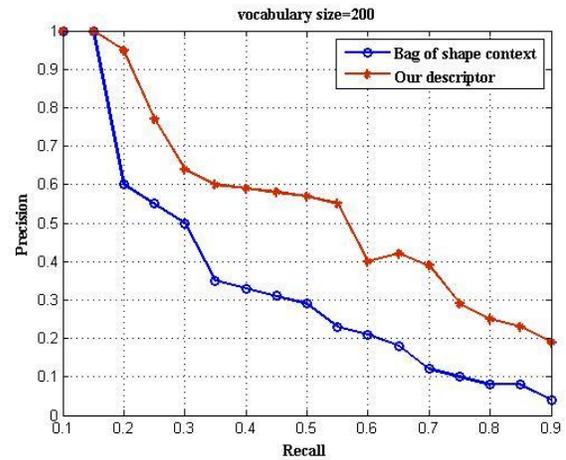

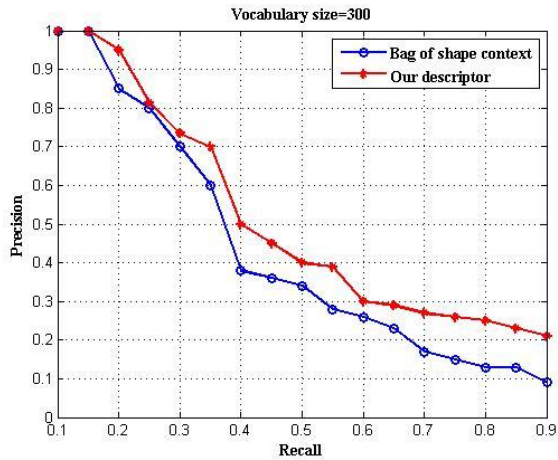

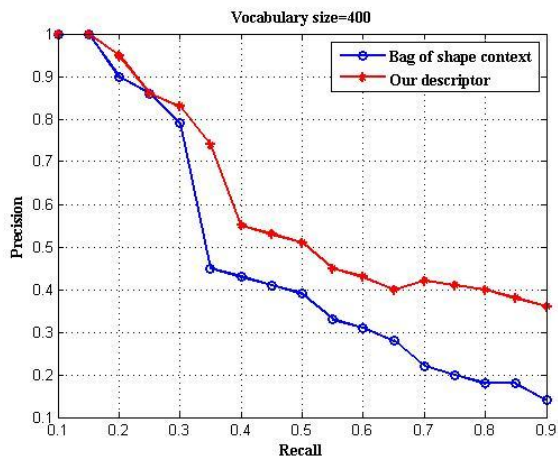

Fig.2: The precision vs. recall curves by our descriptor and bag of shape context for three vocabulary sizes.

several parameters come into play for the quality of result search. The first parameter relate the choie of the number sampled points of the contour. the second parameter is the size of the visual vocabulary. The table 1 summarizes some results of 10-KNN for different values of visual vocabuly sizes and the number of sampled points selectonned from bird contour.

Table 1. average precesion by vocabulary sizes and number of sampled points for bird contors.

| Vocabulary size / Sampled points | Average Precision | | | | | |
|---|---|---|---|---|---|---|
| | k=200 | | k=300 | | k=400 | |
| | 0ur descriptor | Bag of Shape context | 0ur descriptor | Bag of Shape context | 0ur descriptor | Bag of Shape context |
| n=200 | 0.7573 | 0.597 | 0.7655 | 0.674 | 0.77 | 0.699 |
| n=300 | 0.8073 | 0.647 | 0.8155 | 0.724 | 0.82 | 0.749 |
| n=400 | 0.8573 | 0.697 | 0.8655 | 0.774 | 0.87 | 0.799 |
| n=500 | 0.9073 | 0.747 | 0.9155 | 0.824 | 0.92 | 0.849 |
| n=600 | 0.911 | 0.797 | 0.921 | 0.874 | 0.953 | 0.899 |
| n=700 | 0.922 | 0.847 | 0.98 | 0.924 | 0.964 | 0.949 |
| Average Precision | **0.8603** | **0.722** | **0.8771** | **0.799** | **0.8828** | **0.824** |

## 5. Conclusion and future works

In this paper we have developed and tested a method for image retrieval based on the perception of color and shape descriptors. The proposed descriptor exploits the shape context, the color moments together with bag of visual words approach to describe birds. The experimental results are promising.

We plan to extend the work to avoid the severe problems occurring in most existing methods, such as the slow training (e.g., in K-means). In the other axis, we try to avoid determining manually the size of the visual vocabulary by exploiting the learning techniques.